\documentclass[letterpaper, 10 pt, conference]{ieeeconf}  

\IEEEoverridecommandlockouts                              

\usepackage[noadjust]{cite}

\usepackage{graphicx} 
\usepackage{comment}
\usepackage{xcolor}

\usepackage{bm} 
\usepackage{amsmath}
\usepackage{amssymb} 
\usepackage{mathrsfs} 
\usepackage{algorithm} 
\usepackage{mathtools} 

\let\labelindent\relax 
\usepackage{enumitem}
\usepackage{cuted} 
\usepackage{siunitx}

\usepackage{ulem}

\usepackage{kotex}

\usepackage{url}
\usepackage{subfigure}

\usepackage{tikz}

\newcommand\copyrighttext{%
  \footnotesize \textcopyright 2024 IEEE.  Personal use of this material is permitted.  Permission from IEEE must be obtained for all other uses, in any current or future media, including reprinting/republishing this material for advertising or promotional purposes, creating new collective works, for resale or redistribution to servers or lists, or reuse of any copyrighted component of this work in other works.}
\newcommand\copyrightnotice{%
\begin{tikzpicture}[remember picture,overlay]
\node[anchor=south,yshift=10pt] at (current page.south) {\fbox{\parbox{\dimexpr\textwidth-\fboxsep-\fboxrule\relax}{\copyrighttext}}};
\end{tikzpicture}%
}

\title{\LARGE \bf
Autonomous aerial perching and unperching\\using omnidirectional tiltrotor and switching controller
}

\author{Dongjae Lee, Sunwoo Hwang, Jeonghyun Byun, Seung Jae Lee, H. Jin Kim
\thanks{This work was supported in part by Institute of Information \& Communications Technology Planning \& Evaluation (IITP) grant funded by the Korean goverment (MSIT) [NO. 2021-0-01343-004, Artificial Intelligence Graduate School Program (Seoul National University)], in part by Unmanned Vehicles Core Technology Research and Development Program through the National Research Foundation of Korea(NRF) and Unmanned Vehicle Advanced Research Center(UVARC) funded by the Ministry of Science and ICT(NRF-2020M3C1C1A010864), and in part by Basic Science Research Program through the National Research Foundation of Korea(NRF), funded by the Ministry of Education(NRF-2022R1A6A3A13073267).}%
\thanks{Dongjae Lee, Sunwoo Hwang, and Jeonghyun Byun are with the Department of Aerospace Engineering, Seoul National University (SNU), Seoul 08826, South Korea {\tt\small \{ehdwo713, swsw0411, quswjdgus97\}@snu.ac.kr}}%
\thanks{Seung Jae Lee is with the Department of Mechanical System Design Engineering, Seoul National University of Science and Technology (SEOULTECH), Seoul 01811, South Korea {\tt\small seungjae\_lee@seoultech.ac.kr}}%
\thanks{H. Jin Kim is with the Department of Aerospace Engineering, and Interdisciplinary Program in Artificial Intelligence (IPAI), Seoul National University, Seoul, Korea {\tt\small hjinkim@snu.ac.kr}}%
}

\begin{document}

\maketitle
\copyrightnotice
\thispagestyle{empty}
\pagestyle{empty}

\begin{abstract}
Aerial unperching of multirotors has received little attention as opposed to perching that has been investigated to elongate operation time. This study presents a new aerial robot capable of both perching and unperching autonomously on/from a ferromagnetic surface during flight, and a switching controller to avoid rotor saturation and mitigate overshoot during transition between free-flight and perching. To enable stable perching and unperching maneuvers on/from a vertical surface, a lightweight ($\approx$ $1$ \si{kg}), fully actuated tiltrotor that can hover at $90^\circ$ pitch angle is first developed. We design a perching/unperching module composed of a single servomotor and a magnet, which is then mounted on the tiltrotor. A switching controller including exclusive control modes for transitions between free-flight and perching is proposed. Lastly, we propose a simple yet effective strategy to ensure robust perching in the presence of measurement and control errors and avoid collisions with the perching site immediately after unperching. We validate the proposed framework in experiments where the tiltrotor successfully performs perching and unperching on/from a vertical surface during flight. We further show effectiveness of the proposed transition mode in the switching controller by ablation studies where large overshoot and even collision with a perching site occur. To the best of the authors' knowledge, this work presents the first autonomous aerial unperching framework using a fully actuated tiltrotor.
\end{abstract}

\section{Introduction}

Aerial perching can enhance the energy efficiency and flight duration of multirotors whose hovering consumes a significant amount of energy. Such advantage has led to several studies on perching \cite{mao2023robust, ji2022real, lee2023minimally, liu2023hitchhiker}, in particular, on inclined or vertically oriented surfaces.
However, most of these studies have primarily concentrated on the act of perching itself, with little attention given to the process of unperching. Investigating unperching is necessary since, without the capability to autonomously unperch, a multirotor that has perched cannot be reused unless manually retrieved by a human operator. 

\begin{figure}
    \centering
    \includegraphics[width=0.9\linewidth]{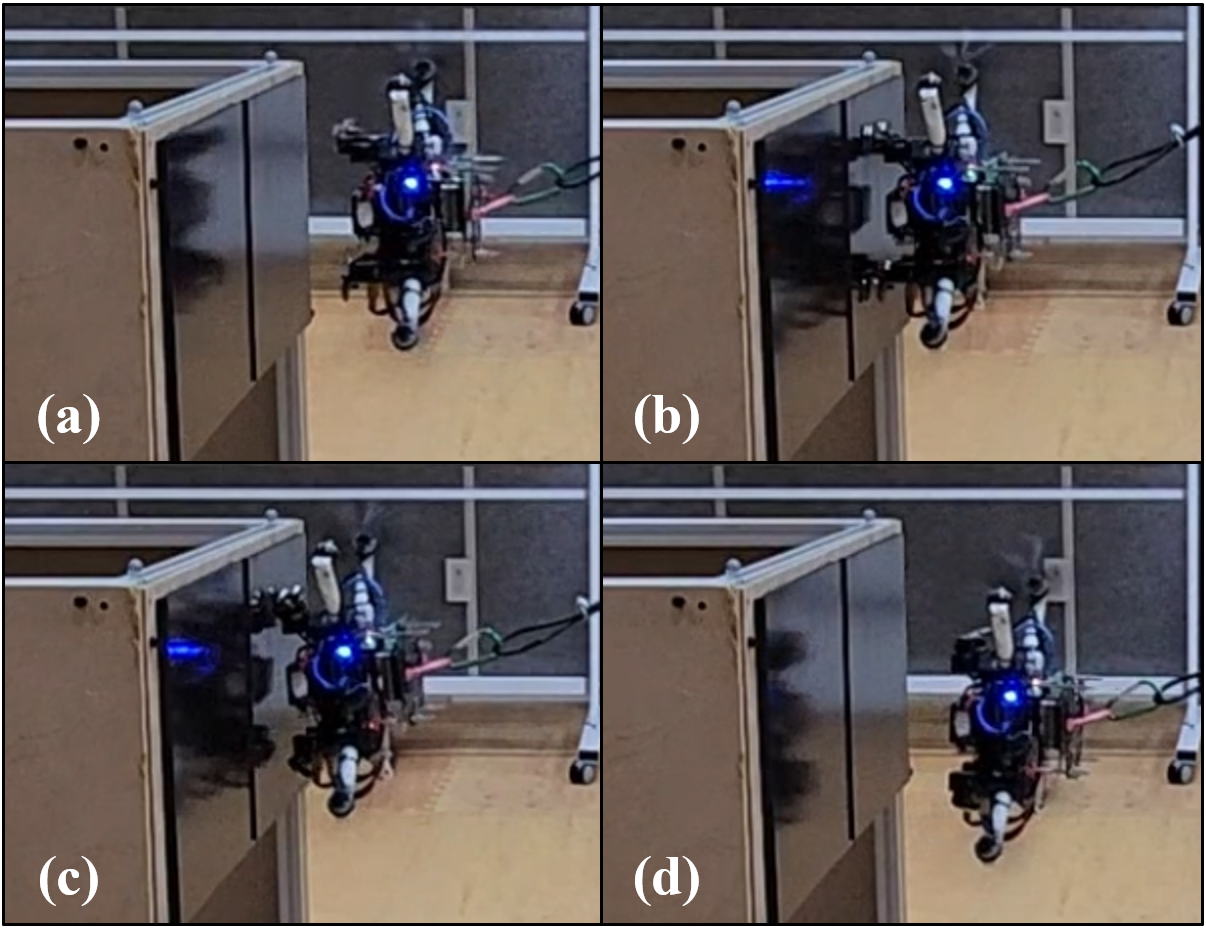}
    \caption{Experimental results showing both autonomous perching ((a) to (b)) and unperching ((c) to (d)) on the fly. Alphabetic order from (a) to (d) indicates the time sequence.}
    \label{fig:thumbnail}
\end{figure}

Since unperching essentially requires perching, we first review research on aerial perching. Most works on aerial perching employed conventional multirotors \cite{ji2022real, mao2023robust, liu2023hitchhiker, yanagimura2014hovering, liu2020adaptive, zhang2021compliant, park2020lightweight, tsukagoshi2015aerial}. However, to perch on a vertical wall or unperch from it, it is necessary to rotate the orientation by nearly $90^\circ$. Conventional multirotors, due to their underactuated nature, inevitably experience translational overshoot during such maneuvers, making them unsuitable for operation in narrow or crowded spaces. Furthermore, to achieve robust alignment between the perching site and the multirotor with respect to control errors and measurement inaccuracy, it is essential to independently control both translation and orientation. Although \cite{lee2023minimally} proposed a tiltrotor that can hover at a non-zero pitch angle to address this issue, it has a limitation of encountering singularity in control allocation when perching on surfaces close to $90^\circ$. Instead of directly tilting the body orientation by $90^\circ$, \cite{tsukagoshi2015aerial} suggests a perching strategy of first attaching to the vertical surface using a suction cup and a valve with zero pitch angle, and then rotating its body to perch on a vertical surface at $90^\circ$ pitch. However, this approach may lack fast response in hovering recovery when encountering sudden detachment or slip due to the conventional multirotor being incapable of hovering at $90^\circ$ pitch angle.

In the aspect of perching mechanism which is an essential factor in aerial perching, most reported mechanisms have not considered unperching motion. Previous works on aerial perching have utilized various materials, such as magnets \cite{ji2022real, lee2023minimally, yanagimura2014hovering}, suction cups \cite{liu2023hitchhiker, liu2020adaptive, tsukagoshi2015aerial}, Velcro \cite{mao2023robust}, grippers \cite{yu2020perching,roderick2021bird,zhang2021compliant}, and electroadhesives \cite{park2020lightweight}. Among those methods, our study extends the research on magnet-based methods. One advantage of using magnets is their capability to perch on various structures with ferromagnetic surfaces. However, most magnet-based perching methods employed in aerial robotics involve firmly attaching the magnet to the multirotor, and thus unperching cannot be executed unless the multirotor's actuation (i.e. thrust force and torque) are utilized. Moreover, using the multirotor's thrust force and torque for unperching may lead to overshoots in position and orientation immediately after unperching, potentially resulting in collision with the perching site and surrounding objects if not carefully controlled.

In addition to the above mentioned problems regarding multirotor platform and perching/unperching mechanism, we address two issues that can be encountered during aerial perching and unperching: rotor saturation during perching and overshoot after unperching. 
As shown in the experimental results of \cite{lee2023minimally}, treating motion constraints as disturbance during perching can result in rotor saturation. Although control techniques like disturbance estimator can enhance control performance during free-flight by compensating model uncertainties including near-wall effect and thrust efficiency change, they may result in adverse effect like rotor saturation during perching. Rotor saturation can cause a significant control error, potentially leading to unintended detachment, and should be mitigated.

Only a few previous research have addressed aerial unperching \cite{zhang2021compliant, yanagimura2014hovering, tsukagoshi2015aerial}, and little attention has been given to reducing undesirable overshoot immediately after unperching. The papers \cite{zhang2021compliant, yanagimura2014hovering, tsukagoshi2015aerial} only consider unperching from \textit{near-zero} roll/pitch angles, and undesirable overshoots in the gravity direction are observed in all papers. Since such overshoot may lead to collisions with the perching site or nearby objects, a more effective method for regulating the motion is needed.

To resolve these problems discussed above, we first present a mechanism capable of both perching and unperching using magnets, specifically designed for application in lightweight, fully actuated multirotors ($\approx$ $1$ \si{kg}). Additionally, we propose a switching controller and motion strategy to prevent rotor saturation and collision with the perching site during the perching and unperching processes. To the best of the authors' knowledge, this paper introduces the first autonomous aerial unperching even at $90^\circ$ pitch angle from a vertical wall using a fully actuated tiltrotor as shown in Fig. \ref{fig:thumbnail}. Contribution of this research can be summarized as follows:
\begin{enumerate}
    \item We propose an aerial robot that is capable of both perching and unperching on a vertical ferromagnetic wall.
    \item We suggest a switching controller and a motion strategy to avoid rotor saturation during perching and overshoot during unperching.
    \item We validate the proposed framework in experiments of autonomous aerial perching and unperching on/from a vertical surface.
\end{enumerate}

\section{Hardware design}

This section presents a hardware design for a multirotor-based aerial robot capable of both perching and unperching on a vertical ferromagnetic surface. The robot is mainly composed of the following two:
\begin{itemize}
    \item a multirotor platform capable of an omnidirectional flight
    \item a mechanism for both perching and unperching using magnet
\end{itemize}

\begin{figure}
    \centering
    \includegraphics[width=0.9\linewidth]{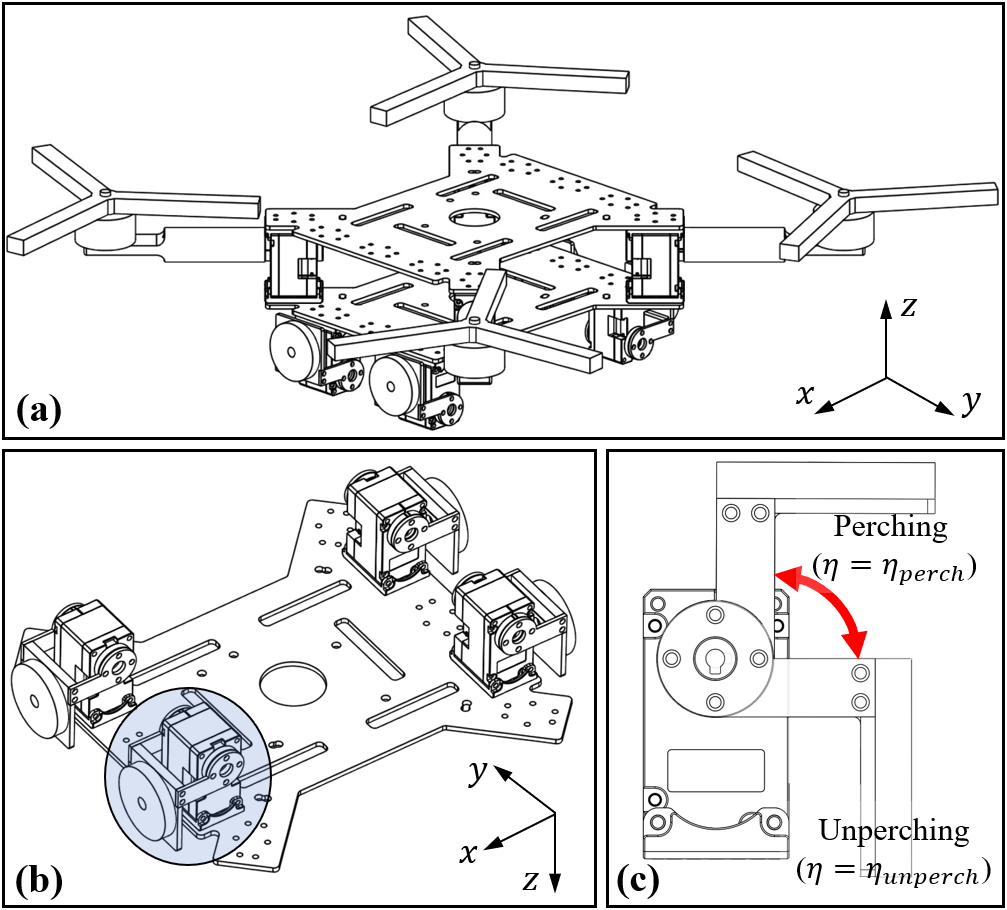}
    \caption{CAD drawings of the tiltrotor (a) and the perching/unperching mechanism (b), (c). The tiltrotor without the perching/unperching mechanism has total of $8$ actuators, including $4$ rotors and $4$ servomotors. The four servomotors rotate the thrust direction of each rotor, enabling full actuation and especially $90^\circ$ pitching while hovering. When unperching, a single perching/unperching module shown in the blue-shaded region in (b) and (c) moves only in a tangential direction to the contact surface, thereby suffering less from adhesion induced by magnets.}
    \label{fig:hardware_mechanism}
\end{figure}

Firstly, to achieve more stable perching and unperching, we developed a lightweight aerial robot capable of hovering at arbitrary orientations. The designed aerial robot, similar to \cite{ryll2015novel} in actuation principle, consists of four rotors and four servomotors where each servomotor independently controls the orientation of its respective rotor. While various fully actuated platforms have been explored, many of them suffer from excessive weight (over 3 kg) \cite{allenspach2020design} or hardware limitations preventing omnidirectional flight \cite{zheng2020tiltdrone,lee2021fully}. In our study, we have developed a relatively lightweight platform (approximately 1 kg, excluding the battery) capable of hovering at arbitrary attitude. The CAD design and the actual drone can be observed in Figs. \ref{fig:hardware_mechanism}(a), \ref{fig:platform}.

Next, to maintain stable perching while minimizing the static torque caused by self-weight during perching, it is essential to keep the distance from the drone's center of mass to the perching site as short as possible. Due to the nature of multirotor drones which require a certain gap between rotors to generate sufficient moment and avoid interference between propellers, the length of the drone in the vertical direction (i.e. $z$-axis direction in Fig. \ref{fig:hardware_mechanism}(a))) can be designed to be much shorter than its length in the forward-backward or lateral directions (i.e. $x,y$-axes directions in Fig. \ref{fig:hardware_mechanism}(a)). Based on this observation, we installed the perching/unperching mechanism on the drone's bottom.

Lastly, as seen in \cite{ji2022real,lee2023minimally}, firmly fixing magnets to the drone body has a significant drawback in that once attached, the magnets are difficult to be detached. To address this issue, we propose a perching and unperching mechanism that utilizes magnets in conjunction with a single servo motor, which we refer to as PS (perching \& unperching servo) to distinguish it from wrench-generating servomotors (WS) used for the tiltrotor's actuation. Leveraging the intuition that magnets exert much weaker forces tangentially than in the surface normal direction, we introduce a rotation-based unperching strategy using the servomotor, as illustrated in Fig. \ref{fig:hardware_mechanism}(c).

We utilized Dynamixel XC330-M288 as a servomotor, weighing approximately 23 g, with a stall torque of 0.93 Nm according to its specifications. Considering the short distance of less than 3 cm from the rotation center of the servo to the magnet, this servo can exert a large tangential force of approximately 30 N, which is sufficient to perform unperching actions when compared to the tiltrotor's total weight of $1.65$ \si{kg}, including tiltrotor platform, perching/unperching mechanism, and battery. To provide sufficient adhesive force during perching, we configured the hardware with four perching and unperching modules attached to the drone's bottom surface, as shown in Fig. \ref{fig:hardware_mechanism}(b).

\section{Switching controller design}

This section presents a switching controller proposed for stable perching and unperching. We control motion during free-flight which corresponds to the flight conditions before perching and after unperching, and we control force during perching considering that motion in all direction is constrained. Then, we suggest a strategy to avoid overshoot and saturation during transition in perching and unperching.

\subsection{System dynamics}
We model system dynamics of the tiltrotor as the following fully actuated rigid body:
\begin{equation}
\begin{aligned}
    \ddot{\bm{p}} &= \frac{1}{m} \bm{R} \bm{f}  -g \bm{b}_3 + \bm{\Delta}_t \\
    \dot{\bm{R}} &= \bm{R} \bm{\omega}^{\wedge} \\
    \dot{\bm{\omega}} &= \bm{J}_b^{-1} (- \bm{\omega} \times \bm{J}_b \bm{\omega} + \bm{\tau}) + \bm{\Delta}_r
\end{aligned}    
\end{equation}
where $\bm{b}_3 = [0;0;1]$, and $g, m \in \mathbb{R}$ and $\bm{J}_b \in \mathbb{R}^{3\times3}$ are gravitational acceleration, total mass and mass moment of inertia of the tiltrotor. $\bm{p}, \bm{\omega} \in \mathbb{R}^3$ are position measured in the world fixed frame and body angular velocity, and $\bm{R} \in \mathsf{SO}(3)$ is the rotation matrix describing the orientation of the tiltrotor. $(\cdot)^{\wedge}$ is an operator mapping a vector in $\mathbb{R}^3$ to a skew-symmetric matrix in $\mathbb{R}^{3\times 3}$. Control inputs are $\bm{f}, \bm{\tau} \in \mathbb{R}^3$, which are forces and torques, and we adopt the control allocation method of \cite{kamel2018voliro}. $\bm{\Delta}_t, \bm{\Delta}_r \in \mathbb{R}^3$ are uncertainty in translation and rotation, for example, contact force and model uncertainty in moment of inertia and center of mass. For multirotors flying with onboard batteries, such uncertainty also includes abrupt thrust efficiency change due to the voltage drop of the batteries.

\begin{figure}
    \centering
    \includegraphics[width=0.8\linewidth]{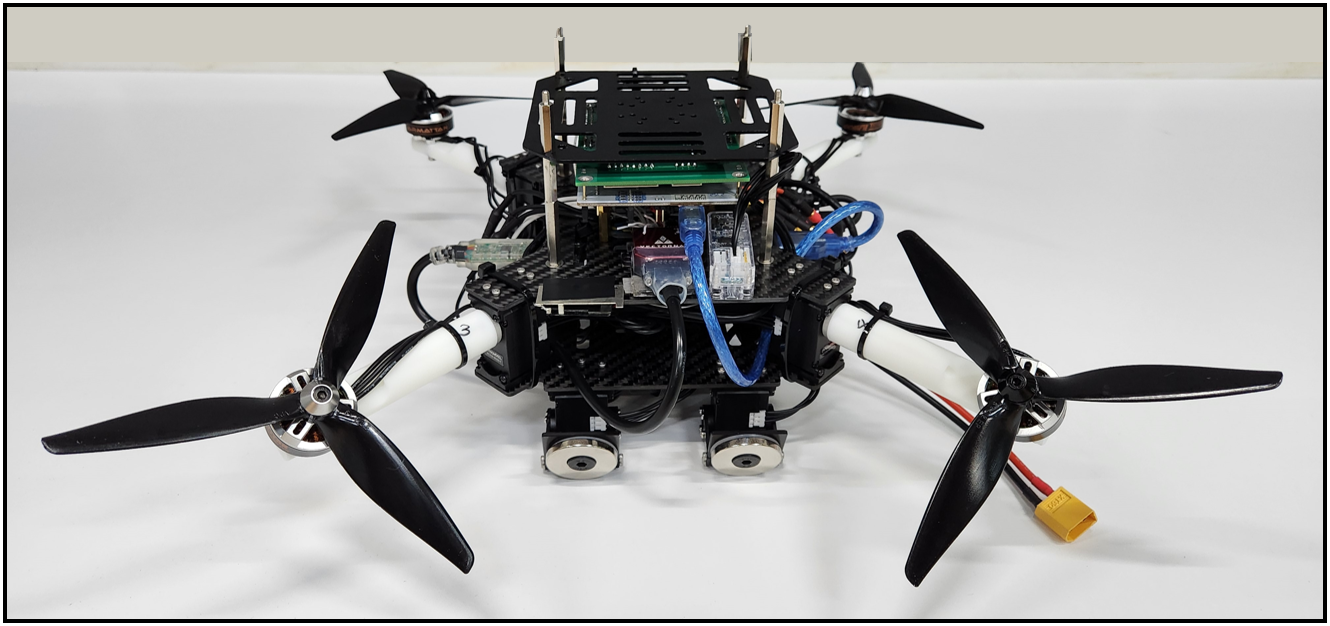}
    \caption{A prototype of the proposed tiltrotor with the perching/unperching modules.}
    \label{fig:platform}
\end{figure}

\subsection{Free-flight controller}
During free-flight, we control motion as follows:
\begin{subequations} \label{eq: ff ctrller}
\begin{align}
    \begin{split} \label{eq: ff ctrller - translation}
        \bm{f} &= \bm{f}_n + \bm{f}_r \\
        \bm{f}_n &= m \bm{R}^\top \left\{g \bm{b}_3 + \bm{K}_{tp} \bm{e}_p + \bm{K}_{td} \dot{\bm{e}}_p + \ddot{\bm{p}}_d \right\} \\
        \bm{f}_r &= -\bm{R}^\top \hat{\bm{\Delta}}_t
    \end{split} \\
    \begin{split} \label{eq: ff ctrller - rotation}
        \bm{\tau} &= \bm{J}_b \left(\bm{K}_{rp} \bm{e}_R + \bm{K}_{rd} \bm{e}_{\omega} + \bm{K}_{ri} \int^t_0 \bm{e}_R dt \right)
    \end{split}
\end{align}
\end{subequations} 
The error terms are defined as $\bm{e}_p = \bm{p}_d - \bm{p}$, $\bm{e}_R = (\text{Log}(\bm{\Psi}))^{\vee}$, $\bm{e}_\omega = \bm{\Psi} \bm{\omega}_d - \bm{\omega}$ with $\bm{\Psi} = \bm{R}^\top \bm{R}_d$. For the error of the rotation matrix, we refer to \cite{yu2016global} instead of \cite{lee2010geometric} since the referred one shows faster convergence rate. $\bm{K}_{tp}, \bm{K}_{td}, \bm{K}_{rp}, \bm{K}_{rd}, \bm{K}_{ri} \in \mathbb{R}^{3\times 3}$ are positive-definite control gains. $\hat{\bm{\Delta}}_t \in \mathbb{R}^3$ is the estimated value through the following momentum-based disturbance estimator \cite{tomic2017external}:
\begin{equation}
\begin{aligned}
    \hat{\bm{\Delta}}_t &= \bm{K}_e \left\{ \bm{p}_m - \bm{p}_m(0) - \int^t_0 (\bm{R} \bm{f} - m g \bm{b}_3 + \hat{\bm{\Delta}}_t) d\tau \right\} \\
    \bm{p}_m &= m \dot{\bm{p}}
\end{aligned}    
\end{equation}
As can be found in \cite[Lemma 1]{alan2022disturbance}, by taking a large estimator gain $\bm{K}_e$, the estimation error $\bm{\Delta}_t - \hat{\bm{\Delta}}_t$ can be maintained to be sufficiently small for all time.

\subsection{Perching controller}
During perching, motion is completely constrained, and the 6-dimensional control inputs are no longer used to control the drone's motion. Instead, they are utilized for controlling the forces and torques applied to the perched surface. In this research, depending on the situation, either zero force/torque can be generated ($\bm{u}_p = \bm{0}_6$), or a force is generated to partially counteract gravity ($f = \rho m g \bm{b}_3$, $\rho \in (0,1)$). 

\subsection{Controller during transitions}

\begin{figure}
    \subfigure[The proposed control law.]{\includegraphics[width=\linewidth]{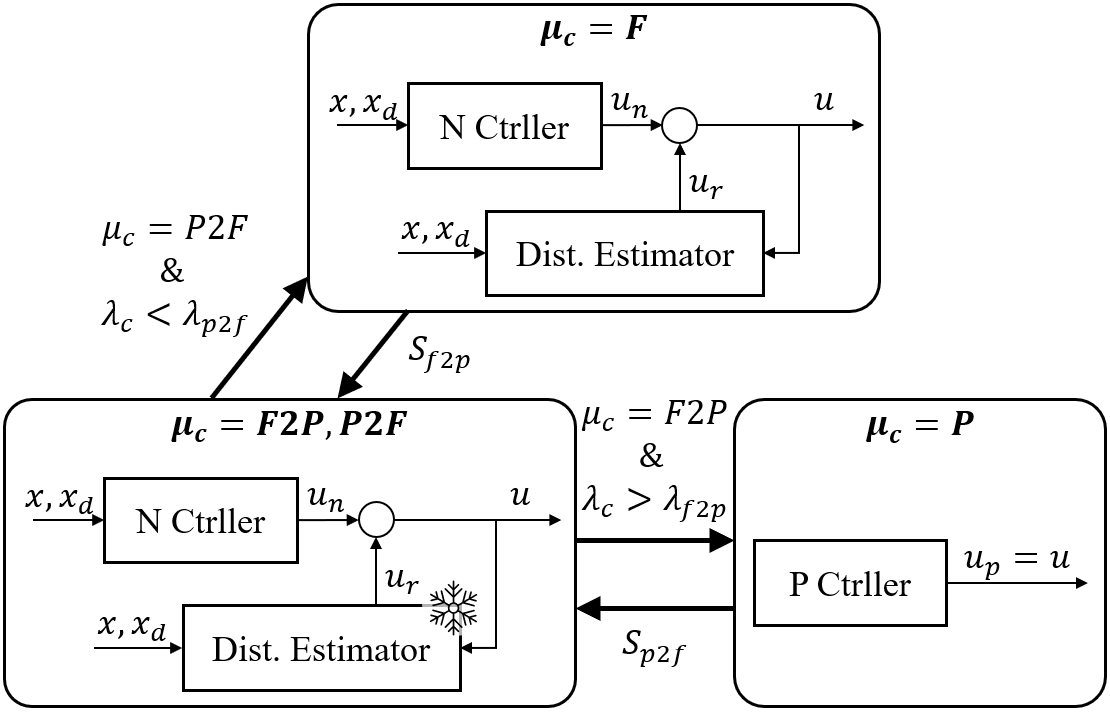}
    \label{fig:ctrl_law}}
    \subfigure[The control law without transition modes.]{\includegraphics[width=\linewidth]{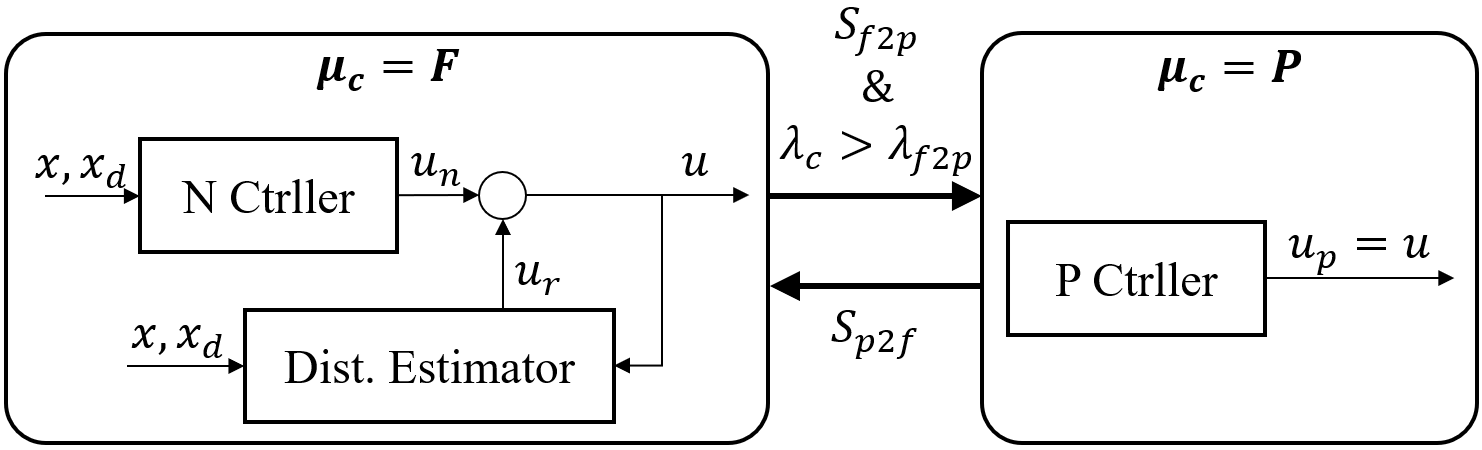}
    \label{fig:ctrl_law_wo_transition}}
    \centering
    \caption{Switching control laws with or without transition modes $F2P, P2F$. $\bm{u}=\bm{u}_n + \bm{u}_r$, $\bm{u}_n = [\bm{f}_n;\bm{\tau}]$, $\bm{u}_r = [\bm{f}_r;\bm{0}_{3}]$}
    \label{fig:ctrl_law_all}
\end{figure}

If transition is not considered separately, only two controllers, i.e. the free-flight controller and the perching controller, would compose the switching controller as shown in Fig. \ref{fig:ctrl_law_wo_transition}. However, without sufficient consideration for transition, there is a possibility of rotor saturation due to blind application of a motion controller during perching, or excessive overshoot due to discontinuous control inputs immediately after unperching.

To address these issues, we propose a separate controller in between the free-flight controller and the perching controller, as depicted in Figure \ref{fig:ctrl_law}. Here, $\mu_c$ represents the mode of the switching controller, with $\mu_c = F$ for free-flight mode, $\mu_c = P$ for perching mode, and $\mu_c = F2P, P2F$ for the transition phases. As seen in the block where $\mu_c = F2P, P2F$ in Fig. \ref{fig:ctrl_law}, the algorithm for disturbance estimation is frozen during all the transition phases, allowing only the nominal controller (N Controller in Fig. \ref{fig:ctrl_law}, and $\bm{f}_n, \bm{\tau}$ in (\ref{eq: ff ctrller - translation})) to operate. Freezing the disturbance estimator during the transition prevents rotor saturation for not considering motion constraints as disturbance, and activating the nominal controller instead of the perching controller during $\mu_c = P2F$ reduces discontinuity when transitioning to the free-flight controller.

The switching signals among control modes are defined using the contact force in the surface-normal direction $\lambda_c \in \mathbb{R}$ and operator inputs $S_{f2p}, S_{p2f} \in \{0, 1\}$. The triggers for transitioning from free-flight to perching and vice versa are assumed to be provided by a human operator. Since it is challenging to precisely determine when perching is successfully achieved, $\lambda_c$ is used to determine the occurrence of perching. If $\lambda_c$ is greater than the threshold $\lambda_{f2p}$, it is deemed that perching has been successfully achieved, and the control mode is switched to perching. Conversely, if $\lambda_c$ decreases below the threshold $\lambda_{p2f}$ while in a perching phase, it is considered that the system is ready to transit to free-flight, and the control mode is switched to free-flight. This switching law is illustrated in Fig. \ref{fig:ctrl_law}.

\section{Perching \& unperching strategy}

\subsection{Perching \& unperching signal generation sequence}
\begin{figure}
    \centering
    \includegraphics[width=\linewidth]{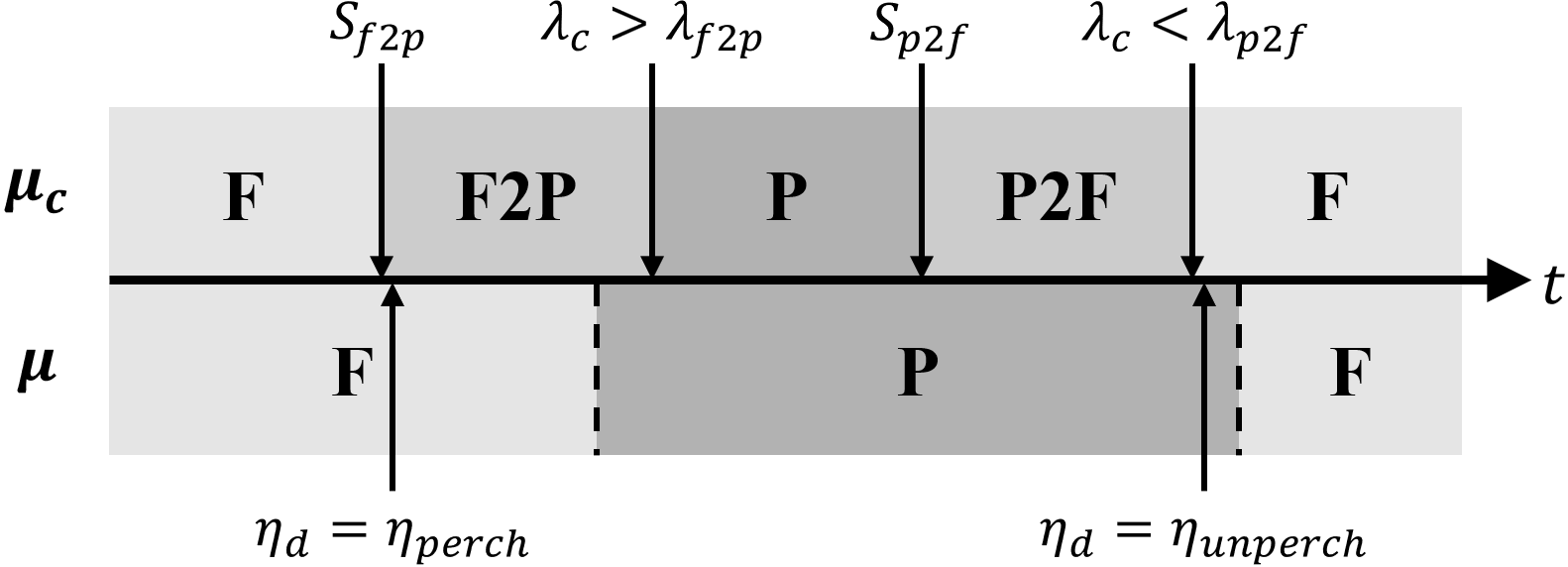}
    \caption{A timeline showing the perching and unperching timings $\eta_d$ with respect to switching signals of the switching controller. $\mu_c$ is the controller mode and $\mu$ is actual status of either perching $P$ or free-flight $F$.}
    \label{fig:mode_switch_seq}
\end{figure}
In the proposed perching/unperching mechanism, as seen in Fig. \ref{fig:hardware_mechanism}(c), perching or unperching is achieved by adjusting the angles of the servomotors PS, to which the magnets are attached ($\eta_d = \eta_{perch}$ for perching and $\eta_d = \eta_{unperch}$ for unperching). It is crucial to determine when to activate the perching/unperching signal $\eta_d$ during the controller mode transition. If $\eta_d$ is set to $\eta_{perch}$ too late during the transition from free-flight to perching, the PSs may move too close to the contact surface, potentially causing collision as they move. Conversely, if $\eta_d$ is switched to $\eta_{unperch}$ too early during the transition from perching to free-flight, the actuators, especially wrench-generating servos (WS), may experience large discontinuity in control inputs. This may cause significant overshoot as WS cannot follow the change quickly enough. This overshoot may lead to collision with the contact surface. To address these issues, we introduce the perching/unperching signal $\eta_d$ for the mechanism immediately after the controller mode $\mu_c$ switches from $F$ to $F2P$ during the transition, as shown in Fig. \ref{fig:mode_switch_seq}. Additionally, when transitioning from perching to free-flight, $\eta_d$ is activated for unperching immediately after $\mu_c$ switches from $P2F$ to $F$.

\subsection{Motion planning}
\begin{figure}
    \centering
    \includegraphics[width=\linewidth]{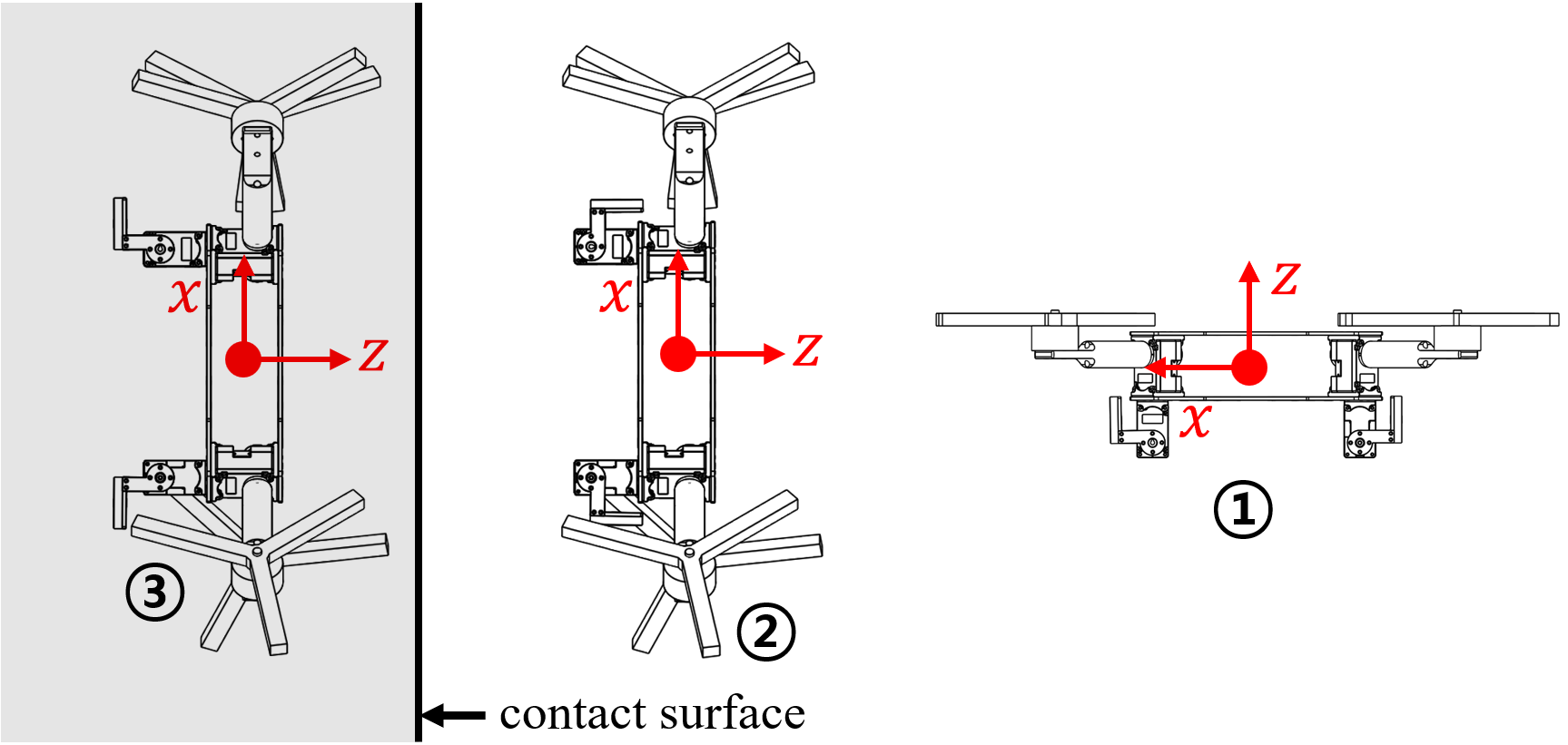}
    \caption{A setpoint generation strategy. A setpoint inside the contact surface as depicted in {\small \textcircled{\footnotesize 3}} is generated to ensure perching, and that apart from the contact surface as in {\small \textcircled{\footnotesize 2}} is given to avoid collision with the perching site during unperching.}
    \label{fig:planning_strategy}
\end{figure}

To minimize the static torque caused by the distance from the contact surface to the center of gravity during perching, a strategy to perch in a 90-degree tilted configuration is adopted. 
Furthermore, for perching and unperching, a trajectory was planned by first setting pose setpoints, as shown in Fig. \ref{fig:planning_strategy} as {\large \textcircled{\small 1}}, {\large \textcircled{\small 2}}, and {\large \textcircled{\small 3}}, and smoothly connecting them. For perching, the setpoint sequence starts with the initial configuration {\large \textcircled{\small 1}} and progresses as {\large \textcircled{\small 1}} $\rightarrow$ {\large \textcircled{\small 2}} $\rightarrow$ {\large \textcircled{\small 3}}, while for unperching, the sequence is set in the reverse order.

When performing perching, a setpoint is placed slightly behind the actual perching site, as shown in Fig. \ref{fig:planning_strategy} {\large \textcircled{\small 3}}. This is to prevent failures in perching due to measurement errors and control performance errors. Additionally, when unperching, a setpoint is placed slightly away from the perching site in the surface normal direction, as seen in {\large \textcircled{\small 2}}. This reduces the net attraction force that must be overcome during unperching and also allows to avoid collision with the perching site immediately after unperching. Trajectory generation between setpoints was done using the method from \cite{brescianini2018computationally}, where minimum jerk and acceleration trajectories in translation and rotation directions respectively were computed in a closed-form when given initial and terminal conditions.

\section{Experiments}
\subsection{Setup}
\begin{figure}
    \centering
    \includegraphics[width=\linewidth]{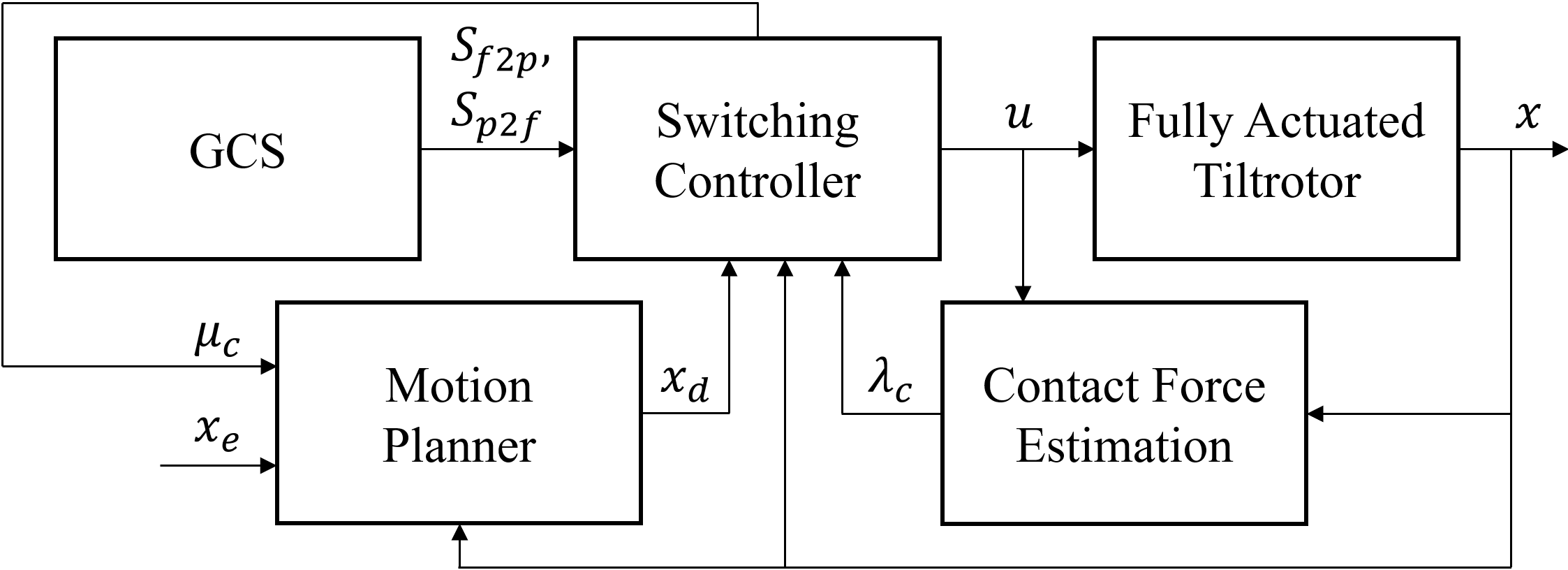}
    \caption{A flow chart showing signal flow during experiments.}
    \label{fig:exp_flowchart}
\end{figure}

The hardware that we built is shown in Fig. \ref{fig:platform}. To actuate the tiltrotor, we used 6-inch propellers with Armattan rotors and controlled the rotor's tilt angle using Dynamixel XC330-M288 servomotors. The magnet-based perching and unperching mechanism also used the same servomotors. For magnets, we used four cylindrical neodymium magnets with the diameter of $30$ \si{mm} and the height of $4$ \si{mm}. We used a $400$ \si{g} $4$S $4200$mAh battery as the power source. Excluding the battery and perching/unperching mechanism, the weight of the tiltrotor itself was approximately 1kg, and the perching and unperching mechanism weighed around 250g. The total weight of the proposed perching/unperching-capable tiltrotor was $1.65$ \si{kg}.

In Fig. \ref{fig:exp_flowchart}, we represented the signal flow among algorithms used during experiments. The ground control station (GCS) operated by a human operator makes decisions about whether to perch or unperch ($S_{f2p}$, $S_{p2f}$). The contact force information used for mode switches between controllers is calculated through a contact force estimator \cite{tomic2017external}. The motion planner generates trajectories based on the controller's mode $\mu_c$ and environmental information ($x_e$) such as the pose of the perching site. For localization, we fused information from the MicroStrain IMU and the motion capture system, Optitrack, to calculate the tiltrotor's odometry at 400 Hz. All algorithms except for the GCS were executed on an onboard Intel NUC computer. We used Ubuntu 20.04 and ROS for software support.

In this study, we conducted perching and unperching experiments on a ferromagnetic vertical wall to validate the feasibility of the proposed mechanism and controller. We also conducted comparative experiments to demonstrate the superiority of the proposed control method when not applied.

\subsection{Results}
\subsubsection{Without the proposed method}
\begin{figure}
    \centering
    \includegraphics[width=\linewidth]{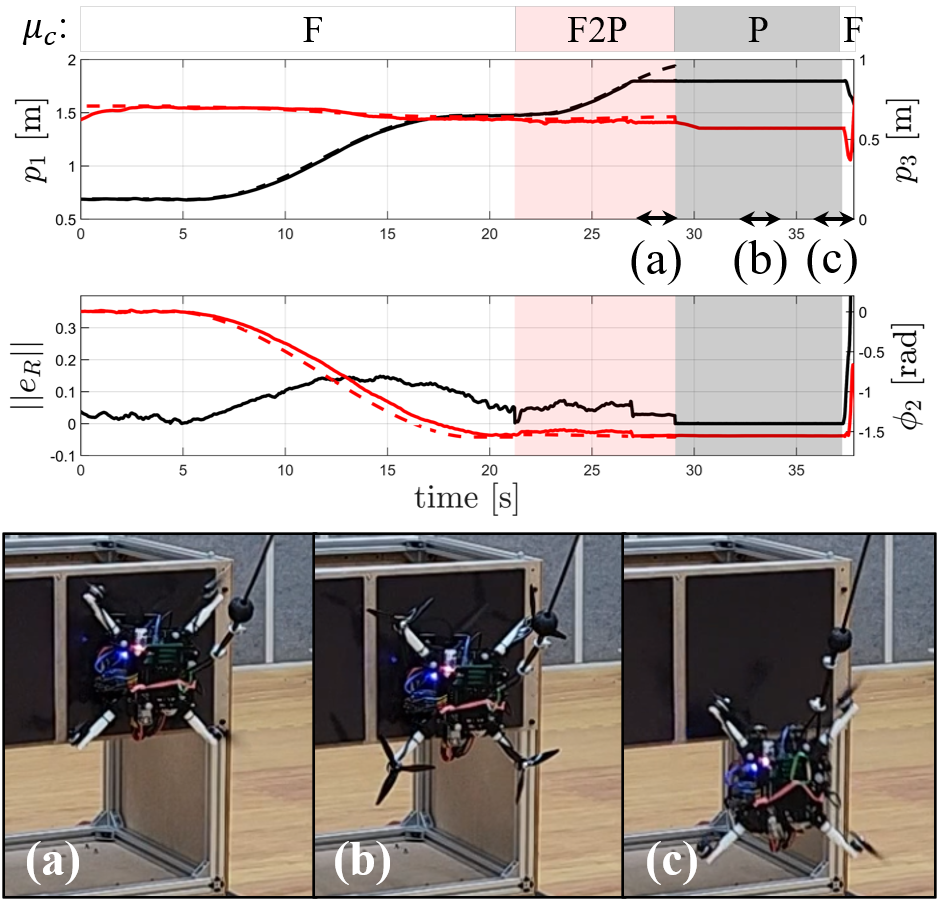}
    \caption{The first experimental results without the proposed method showing a time history of position $p$, pitch angle $\phi_2$, and orientation error $\lVert e_R \rVert$. The shaded regions indicate the controller mode during that period. The captured images show the tiltrotor's status during corresponding periods (a), (b), and (c) in the graph.}
    \label{fig:result_overshoot}
\end{figure}

\begin{figure}
    \centering
    \includegraphics[width=\linewidth]{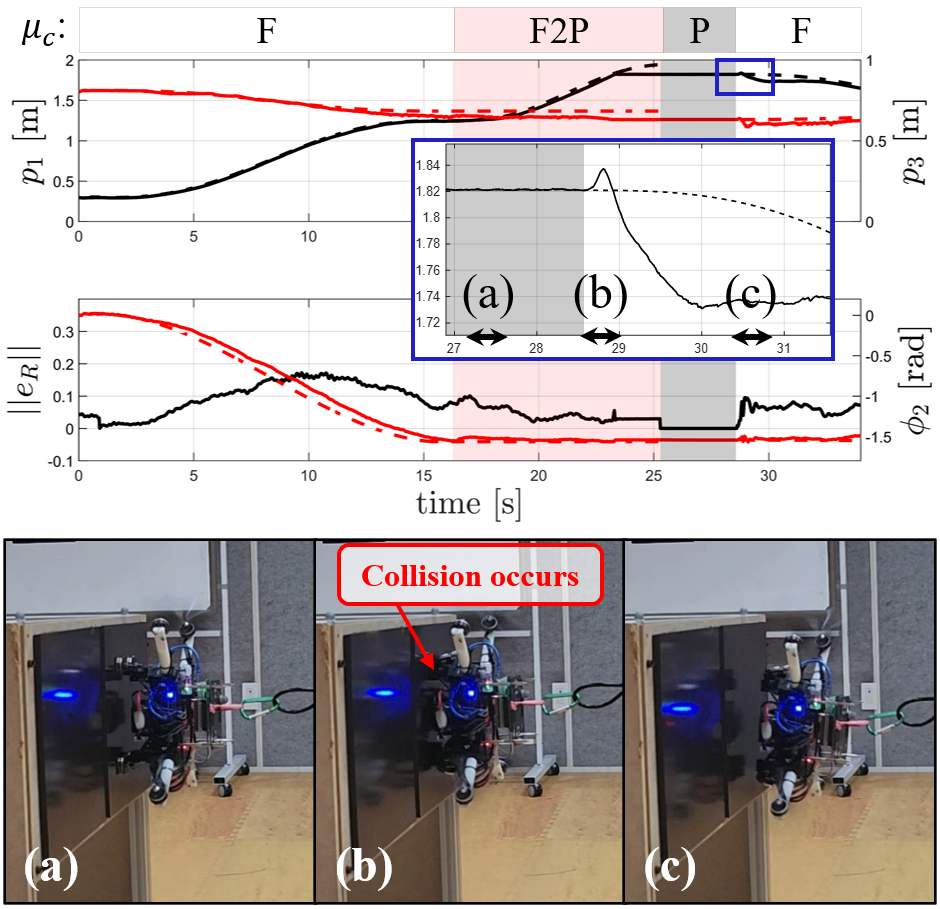}
    \caption{The second experimental results without the proposed method showing a time history of position $p$, pitch angle $\phi_2$, and orientation error $\lVert e_R \rVert$. The shaded regions indicate the controller mode during that period. The captured images show the tiltrotor's status during corresponding periods (a), (b), and (c) in the graph.}
    \label{fig:result_collision}
\end{figure}

In this study, we conducted experiments on cases without the $P2F$ mode among the proposed techniques to confirm the necessity of the proposed $P2F$ controller. In cases without the $P2F$ mode, as shown in Fig. \ref{fig:ctrl_law_wo_transition}, the system transitions directly from the $P$ mode to the $F$ mode. We experimented with two scenarios: 1) when $\bm{f} = \bm{0}_3$ and $\bm{\tau} = \bm{0}_3$ in the $P$ mode, and 2) when $\bm{f} = 0.5 m g \bm{b}_3$ and $\bm{\tau} = \bm{0}_3$ in the $P$ mode. The experimental results can be observed in Figs. \ref{fig:result_overshoot} and \ref{fig:result_collision}. To comply with page limits, we display only the x and z directional positions and pitch angle, along with the magnitude of orientation errors $\lVert e_R \rVert$, showing the most pronounced changes during the experiments.

The results of the first case can be seen in Fig. \ref{fig:result_overshoot}. As shown in (c), there was a sudden drop in the downward direction immediately after unperching, which posed a risk of collision with the ground. To prevent this, the controller was forcibly turned off, and a safety rope was activated. The graphs represent the values only before pulling the safety rope. As anticipated, the abrupt transition from the $P$ mode to the $F$ mode resulted in a significant jump in control inputs, especially because the $WS$ did not track it rapidly enough, leading to a substantial overshoot.

The results of the second case, where control inputs in the $P$ mode were set to $\bm{f} = 0.5 m g \bm{b}_3$, can be found in Fig. \ref{fig:result_collision}. In this case, since $WS$ were already oriented to offset gravity in the $P$ mode, the transition from the $P$ mode to the $F$ mode did not result in significant changes in the servomotor angles, causing small overshoot in the z-axis direction. However, unlike the previous $P2F$ mode where setpoints were provided in the detachment direction and a motion controller was active to follow them, in this case, the motion controller was not activated during perching. Therefore, control inputs for detaching were not generated in advance, and collision with the perching site immediately after unperching occurred, as shown in Fig. \ref{fig:result_collision}(b).

\subsubsection{With the proposed method}

\begin{figure}
    \centering
    \includegraphics[width=\linewidth]{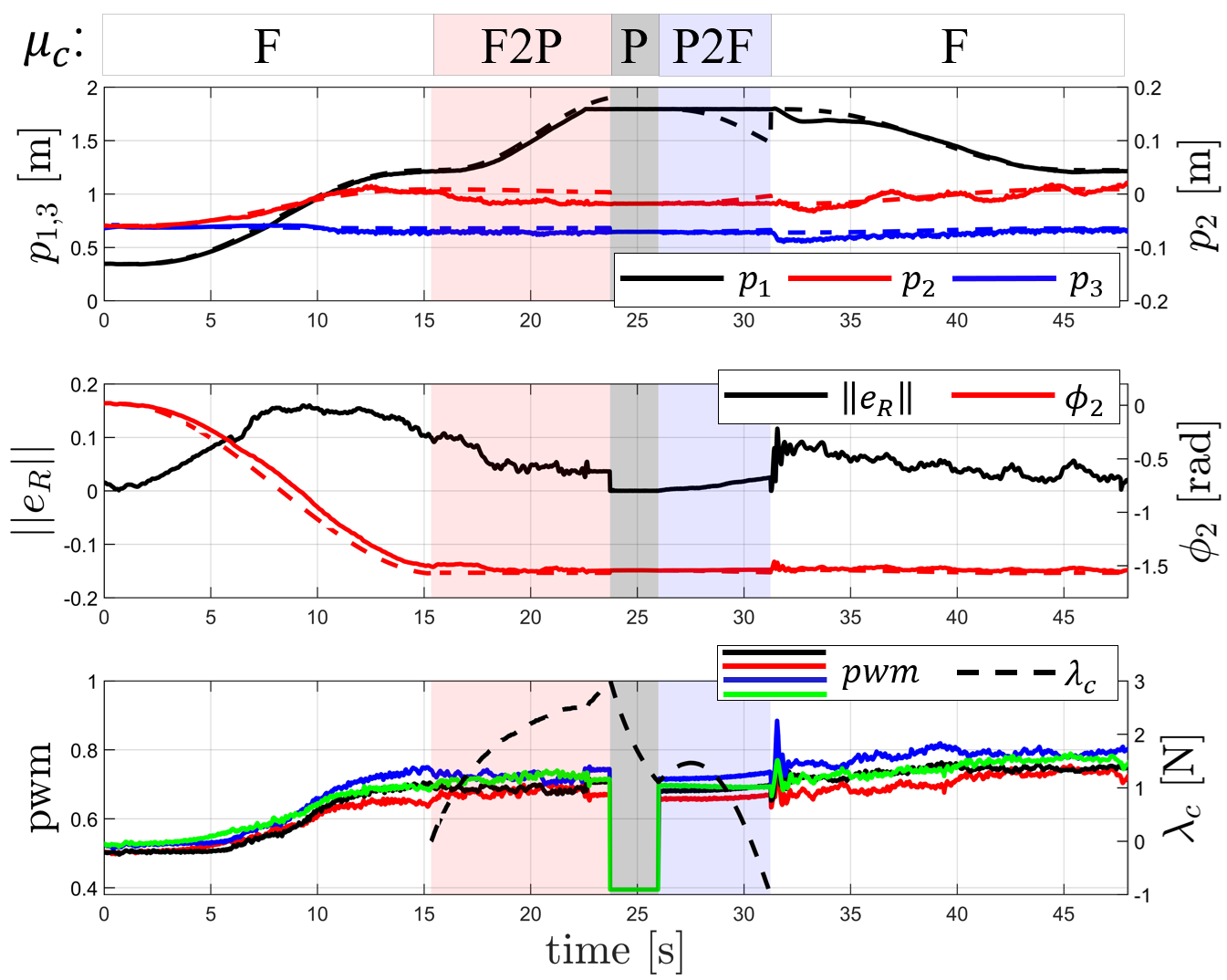}
    \caption{Experimental results with the proposed method showing a time history of position $p$, pitch angle $\phi_2$, orientation error $\lVert e_R \rVert$, PWM, and estimated contact force $\lambda_c$. The shaded regions indicate a controller mode during that period.}
    \label{fig:result_success}
\end{figure}

To validate the proposed method, we conducted perching and unperching experiments on a virtual wall with a ferromagnetic surface. Information about the perching site was obtained using Optitrack. The scenario of this experiment is as follows: (1) hover at 0 degrees roll and pitch, (2) transit to a $90^\circ$ pitch angle, (3) approach the vertical wall and perch, (4) maintain perching for a while, and (5) unperch and finally return to a stationary hover. The result of this experiment can be seen in Fig. \ref{fig:thumbnail}, and more detailed information can be found in the attached video. Since this experiment was conducted to validate perching and unperching capabilities, we only aimed to maintain the perch during perching. Further research on interaction control after perching is considered as future work. To assist in maintaining the perch during perching, we set $\bm{f} = 0.5 m g \bm{b}_3$ and $\bm{\tau} = \bm{0}_3$.

Data collected during the experiment, including position $p$, pitch angle $\phi_2$, magnitude of the orientation error $\lVert e_R \rVert$, PWM, and estimated contact force $\lambda_c$, can be seen in Fig. \ref{fig:result_success}. In the top two figures, the solid lines represent measurement data, while dashed lines represent the desired trajectory of the corresponding solid lines in the same color. Contact force estimation was not active during free flight, so data was measured only during the modes other than free flight. The shaded regions indicate the controller's mode, where white, red, black, and blue correspond to $F$, $F2P$, $P$, and $P2F$ respectively, as indicated in the top right of the figure.

\section{Conclusion}
In this paper, we proposed an aerial robot design capable of in-flight perching and unperching, along with a switching controller tailored for these tasks. First, we developed a lightweight, fully actuated tiltrotor capable of hovering even at $90^\circ$ pitching to perform stable perching and unperching on vertical walls, and designed a perching/unperching module. The switching control law was devised to address challenges such as rotor saturation during perching and excessive overshoot immediately after unperching. Moreover, we proposed a simple yet effective strategy to ensure robust perching execution in the presence of measurement and control errors and avoid collisions with the perching site after unperching. We validated our proposed methods through real-world experiments, successfully demonstrating stable in-flight perching and unperching maneuvers. Additionally, an ablation study involving one of the proposed switching control modes confirmed effectiveness of the proposed switching controller in reducing overshoot. Future work will involve analyses and comparative experiments in a wider range of settings to further evaluate our proposed methodologies.

\addtolength{\textheight}{-12cm}   

\normalem 


\begin{thebibliography}{10}
\providecommand{\url}[1]{#1}
\csname url@rmstyle\endcsname
\providecommand{\newblock}{\relax}
\providecommand{\bibinfo}[2]{#2}
\providecommand\BIBentrySTDinterwordspacing{\spaceskip=0pt\relax}
\providecommand\BIBentryALTinterwordstretchfactor{4}
\providecommand\BIBentryALTinterwordspacing{\spaceskip=\fontdimen2\font plus
\BIBentryALTinterwordstretchfactor\fontdimen3\font minus
  \fontdimen4\font\relax}
\providecommand\BIBforeignlanguage[2]{{%
\expandafter\ifx\csname l@#1\endcsname\relax
\typeout{** WARNING: IEEEtran.bst: No hyphenation pattern has been}%
\typeout{** loaded for the language `#1'. Using the pattern for}%
\typeout{** the default language instead.}%
\else
\language=\csname l@#1\endcsname
\fi
#2}}

\bibitem{mao2023robust}
J.~Mao, S.~Nogar, C.~M. Kroninger, and G.~Loianno, ``Robust active visual
  perching with quadrotors on inclined surfaces,'' \emph{IEEE Transactions on
  Robotics}, 2023.

\bibitem{ji2022real}
J.~Ji, T.~Yang, C.~Xu, and F.~Gao, ``Real-time trajectory planning for aerial
  perching,'' in \emph{2022 IEEE/RSJ International Conference on Intelligent
  Robots and Systems (IROS)}.\hskip 1em plus 0.5em minus 0.4em\relax IEEE,
  2022, pp. 10\,516--10\,522.

\bibitem{lee2023minimally}
D.~Lee, S.~Hwang, C.~Kim, S.~J. Lee, and H.~J. Kim, ``Minimally actuated
  tiltrotor for perching and normal force exertion,'' in \emph{2023 IEEE/RSJ
  International Conference on Intelligent Robots and Systems (IROS)}.\hskip 1em
  plus 0.5em minus 0.4em\relax IEEE, 2023, pp. 5027--5033.

\bibitem{liu2023hitchhiker}
S.~Liu, Z.~Wang, X.~Sheng, and W.~Dong, ``Hitchhiker: A quadrotor aggressively
  perching on a moving inclined surface using compliant suction cup gripper,''
  \emph{IEEE Transactions on Automation Science and Engineering}, 2023.

\bibitem{yanagimura2014hovering}
K.~Yanagimura, K.~Ohno, Y.~Okada, E.~Takeuchi, and S.~Tadokoro, ``Hovering of
  mav by using magnetic adhesion and winch mechanisms,'' in \emph{2014 IEEE
  International Conference on Robotics and Automation (ICRA)}.\hskip 1em plus
  0.5em minus 0.4em\relax IEEE, 2014, pp. 6250--6257.

\bibitem{liu2020adaptive}
S.~Liu, W.~Dong, Z.~Ma, and X.~Sheng, ``Adaptive aerial grasping and perching
  with dual elasticity combined suction cup,'' \emph{IEEE Robotics and
  Automation Letters}, vol.~5, no.~3, pp. 4766--4773, 2020.

\bibitem{zhang2021compliant}
H.~Zhang, E.~Lerner, B.~Cheng, and J.~Zhao, ``Compliant bistable grippers
  enable passive perching for micro aerial vehicles,'' \emph{IEEE/ASME
  Transactions on Mechatronics}, vol.~26, no.~5, pp. 2316--2326, 2021.

\bibitem{park2020lightweight}
S.~Park, D.~S. Drew, S.~Follmer, and J.~Rivas-Davila, ``Lightweight high
  voltage generator for untethered electroadhesive perching of micro air
  vehicles,'' \emph{IEEE Robotics and Automation Letters}, vol.~5, no.~3, pp.
  4485--4492, 2020.

\bibitem{tsukagoshi2015aerial}
H.~Tsukagoshi, M.~Watanabe, T.~Hamada, D.~Ashlih, and R.~Iizuka, ``Aerial
  manipulator with perching and door-opening capability,'' in \emph{2015 IEEE
  International Conference on Robotics and Automation (ICRA)}.\hskip 1em plus
  0.5em minus 0.4em\relax IEEE, 2015, pp. 4663--4668.

\bibitem{yu2020perching}
P.~Yu, G.~Chamitoff, and K.~Wong, ``Perching upside down with bi-directional
  thrust quadrotor,'' in \emph{2020 International Conference on Unmanned
  Aircraft Systems (ICUAS)}.\hskip 1em plus 0.5em minus 0.4em\relax IEEE, 2020,
  pp. 1697--1703.

\bibitem{roderick2021bird}
W.~R. Roderick, M.~R. Cutkosky, and D.~Lentink, ``Bird-inspired dynamic
  grasping and perching in arboreal environments,'' \emph{Science Robotics},
  vol.~6, no.~61, p. eabj7562, 2021.

\bibitem{ryll2015novel}
M.~Ryll, H.~H. B{\"u}lthoff, and P.~R. Giordano, ``A novel overactuated
  quadrotor unmanned aerial vehicle: Modeling, control, and experimental
  validation,'' \emph{IEEE Trans. Control Syst. Technol.}, vol.~23, no.~2, pp.
  540--556, 2015.

\bibitem{allenspach2020design}
M.~Allenspach, K.~Bodie, M.~Brunner, L.~Rinsoz, Z.~Taylor, M.~Kamel,
  R.~Siegwart, and J.~Nieto, ``Design and optimal control of a tiltrotor
  micro-aerial vehicle for efficient omnidirectional flight,'' \emph{Int. J.
  Rob. Res.}, vol.~39, no. 10-11, pp. 1305--1325, 2020.

\bibitem{zheng2020tiltdrone}
P.~Zheng, X.~Tan, B.~B. Kocer, E.~Yang, and M.~Kovac, ``Tiltdrone: A
  fully-actuated tilting quadrotor platform,'' \emph{IEEE Robotics and
  Automation Letters}, vol.~5, no.~4, pp. 6845--6852, 2020.

\bibitem{lee2021fully}
S.~J. Lee, D.~Lee, J.~Kim, D.~Kim, I.~Jang, and H.~J. Kim, ``Fully actuated
  autonomous flight of thruster-tilting multirotor,'' \emph{IEEE/ASME
  Transactions on Mechatronics}, vol.~26, no.~2, pp. 765--776, 2021.

\bibitem{kamel2018voliro}
M.~Kamel, S.~Verling, O.~Elkhatib, C.~Sprecher, P.~Wulkop, Z.~Taylor,
  R.~Siegwart, and I.~Gilitschenski, ``The voliro omniorientational hexacopter:
  An agile and maneuverable tiltable-rotor aerial vehicle,'' \emph{IEEE
  Robotics \& Automation Magazine}, vol.~25, no.~4, pp. 34--44, 2018.

\bibitem{yu2016global}
Y.~Yu and X.~Ding, ``A global tracking controller for underactuated aerial
  vehicles: design, analysis, and experimental tests on quadrotor,''
  \emph{IEEE/ASME Transactions on Mechatronics}, vol.~21, no.~5, pp.
  2499--2511, 2016.

\bibitem{lee2010geometric}
T.~Lee, M.~Leok, and N.~H. McClamroch, ``Geometric tracking control of a
  quadrotor uav on se (3),'' in \emph{49th IEEE conference on decision and
  control (CDC)}.\hskip 1em plus 0.5em minus 0.4em\relax IEEE, 2010, pp.
  5420--5425.

\bibitem{tomic2017external}
T.~Tomi{\'c}, C.~Ott, and S.~Haddadin, ``External wrench estimation, collision
  detection, and reflex reaction for flying robots,'' \emph{IEEE Transactions
  on Robotics}, vol.~33, no.~6, pp. 1467--1482, 2017.

\bibitem{alan2022disturbance}
A.~Alan, T.~G. Molnar, E.~Da{\c{s}}, A.~D. Ames, and G.~Orosz, ``Disturbance
  observers for robust safety-critical control with control barrier
  functions,'' \emph{IEEE Control Systems Letters}, vol.~7, pp. 1123--1128,
  2022.

\bibitem{brescianini2018computationally}
D.~Brescianini and R.~D’Andrea, ``Computationally efficient trajectory
  generation for fully actuated multirotor vehicles,'' \emph{IEEE Transactions
  on Robotics}, vol.~34, no.~3, pp. 555--571, 2018.

\end{thebibliography}

\end{document}